\documentclass[10pt,conference]{IEEEtran}
\IEEEoverridecommandlockouts

\usepackage{cite}
\usepackage{amsmath,amssymb,amsfonts}
\usepackage{algorithmic}
\usepackage{graphicx}
\usepackage{textcomp}
\usepackage{xcolor}
\usepackage{fancyhdr}
\def\BibTeX{{\rm B\kern-.05em{\sc i\kern-.025em b}\kern-.08em
    T\kern-.1667em\lower.7ex\hbox{E}\kern-.125emX}}

 \fancypagestyle{firstpage}{%
 \fancyhf{}%
\fancyhead[L]{2025 IEEE Military Communications Conference}

  \fancyfoot[L]{%
\normalsize{979-8-3503-8359-1/24/\$31.00 ~\copyright2024 IEEE}
}%
 }

\usepackage[top=0.75in, bottom=1.037in, left=0.7in, right=0.7in]{geometry}

\begin{document}

\title{
    {Collateral Damage Assessment Model for AI System Target Engagement in Military Operations}
}

\author{
     \IEEEauthorblockN{Clara Maathuis\IEEEauthorrefmark{1}, Kasper Cools\IEEEauthorrefmark{2}\IEEEauthorrefmark{3}
    }
    
    \IEEEauthorblockA{
        \IEEEauthorrefmark{1}Open University of the Netherlands\\
        \IEEEauthorrefmark{2}Royal Military Academy, Belgium\\
        \IEEEauthorrefmark{3}Vrije Universiteit Brussel, Belgium\\
        \IEEEauthorrefmark{1}clara.maathuis@ou.nl, 
                 \IEEEauthorrefmark{2}kasper.cools@mil.be 
    } 
}

\maketitle
\thispagestyle{firstpage}

\begin{abstract}
In an era where AI (Artificial Intelligence) systems play an increasing role in the battlefield, ensuring responsible targeting demands rigorous assessment of potential collateral effects. In this context, a novel collateral damage assessment model for target engagement of AI systems in military operations is introduced. The model integrates temporal, spatial, and force dimensions within a unified Knowledge Representation and Reasoning (KRR) architecture following a design science methodological approach. Its layered structure captures the categories and architectural components of the AI systems to be engaged together with corresponding engaging vectors and contextual aspects. At the same time, spreading, severity, likelihood, and evaluation metrics are considered in order to provide a clear representation enhanced by transparent reasoning mechanisms. Further, the model is demonstrated and evaluated through instantiation which serves as a basis for further dedicated efforts that aim at building responsible and trustworthy intelligent systems for assessing the effects produced by engaging AI systems in military operations. 
\end{abstract}


\section{Introduction}
Artificial intelligence (AI) systems and AI-based capabilities, such as AI decision support systems (AI-DSS) and AI-enabled operational tools, are increasingly used in the military domain \cite{I1, I13}. These systems are integrated across intelligence gathering and surveillance to logistics, planning, and target acquisition \cite{I2, I3}. The development and deployment of predictive AI systems enables rapid identification and countering of threats like hypersonic missiles and intelligent cyber weapons, while AI-driven decision support capabilities demonstrate advances in optimising intelligence collection, target prioritisation, and resource allocation under complex and time-sensitive conditions \cite{I4, I5, I6, I14}. Such systems enhance precision, increase operational tempo, and support human decision-makers, reflecting the transformative impact on military strategy, decision cycles, and battlefield coordination.

Through their integration into military workflows, AI systems themselves become potential targets in military operations. Disrupting or degrading an adversary's AI-enabled infrastructure, whether data-driven, knowledge-based, or neuro-symbolic, may yield significant operational advantage \cite{I7, US3, I15, I16}. This development accentuates the importance of rigorous assessment frameworks when planning and executing attacks on such targets in a legal and responsible way. Military operations must account for both mission-related aspects and legal obligations arising under international humanitarian law, such as the principles of distinction and proportionality \cite{I8, I9, US1}. Building AI solutions in target selection, engagement, and collateral damage assessment amplifies the complexity and uncertainty of compliance, making intelligent and adaptive assessment models essential.

Collateral damage is incidental, unintended harm to civilians and civilian objects during attacks on lawful military objectives. This includes civilian casualties and property damage, which must not be excessive relative to the anticipated military advantage, as stipulated in Additional Protocol I Articles 51(5)(b) and 57(2)(a)(iii) \cite{CD1, CD2}. Civilian harm must be both unintended and proportional; excessive anticipated harm renders attacks unlawful \cite{CD3}. Commanders must take all feasible precautions in planning and execution, including suspension if circumstances change \cite{CD4}. For emerging technologies, collateral damage extends to civilian infrastructure and data integrity affected incidentally \cite{CD5}.

While collateral damage assessment methods exist in kinetic and cyber settings \cite{I10, I11, I12, US2}, corresponding methods for AI system engagement are needed. This research develops a collateral damage assessment model for AI system target engagement in military operations, accounting for temporal, spatial, and force dimensions as well as severity and likelihood of unintended effects on civilians and civilian objects. The model is developed as a computational ontology following a Design Science Research (DSR) methodological approach \cite{DSR} respecting Knowledge Representation and Reasoning (KRR) principles \cite{KRR}. This allows capturing the full spectrum of AI systems-data-driven, knowledge-driven, and neuro-symbolic models-along with their key system components, operational relationships, and civilian linkages. This formalism supports granular annotation of technical and human-centric attributes, enabling encoding of datasets, models, inference engines, civilian infrastructure, cultural contexts, and their dependencies.

This research advances both military and AI domains by providing an adaptive, transparent computational model for collateral damage assessment in AI-driven warfare. It introduces a structured approach bridging kinetic and non-kinetic elements while embedding legal, ethical, and social considerations into responsible AI targeting decisions. The model's integration of system architecture, explainability, validation, and risk mitigation establishes a foundation for responsible AI solutions, particularly dual-use systems, contributing to perceiving AI systems as socio-technical systems in military contexts.

This article is structured as follows. Section \ref{section:related-work} discusses related studies. Section \ref{section:research-methodology} provides an overview of the research methodology. Section \ref{section:modeldesign} presents the model design and development. Section \ref{section:modelevaluation} instantiates the model on a use case. Section \ref{section:conclusion} discusses concluding remarks and future research perspectives.

\section{Related Work\label{section:related-work}}
The integration of AI systems into military operations has generated extensive research on their strategic, ethical, and legal implications \cite{LavenderGaza2025, RobertsAnastasiaVenables21}. AI decision-support and operational tools are now deployed in surveillance, logistics, and target acquisition, enhancing operational tempo and precision in high-stakes environments \cite{I1, I2, I3, I4, I5, I6}. Collateral damage assessment (CDA) models have traditionally focused on kinetic operations, using probabilistic and simulation-based methods \cite{I10, I11}. In the cyber domain, studies assess unintended impacts on civilian infrastructure and data integrity \cite{I12, US2, OCO2025WLCheng}. However, non-kinetic targeting of AI systems-such as disrupting data pipelines, degrading inference engines, or manipulating model behaviour-remains underexplored despite its growing relevance. Doctrinal analyses define non-kinetic targeting as the use of military and non-military means (information operations, lawfare, cyber actions, disinformation, espionage) to influence adversaries without physical force \cite{Ducheine2016, ratiu2024}, but conceptual and computational frameworks to assess its effects-especially on AI systems-are scarce. Legal and ethical scholarship has examined targeting under IHL principles like distinction and proportionality, noting challenges from dual-use and opaque AI systems \cite{I8, I9, US1}. Yet few computational models capture temporal, spatial, and force dimensions in non-kinetic contexts. Ontological approaches in Knowledge Representation and Reasoning (KRR) offer potential for modeling complexity, explainability, and risk in AI \cite{KRR}, but applications to military CDA are limited and often neglect socio-technical dependencies. Growing AI use in military decision-making raises concerns over transparency, robustness, and susceptibility to manipulation \cite{svenmarck2018possibilities, wood_explainable_2024}, making structured, trustworthy assessment models essential-particularly where subtle, non-kinetic disruptions can have significant consequences. This work addresses these gaps with a layered KRR-based model unifying temporal, spatial, and force dimensions for responsible CDA in non-kinetic engagements of AI systems.

\section{Research Methodology\label{section:research-methodology}}
This research aims to build a computational model for representing and reasoning on the meaning and assessment of collateral damage as it follows the target engagement of AI systems in military operations. It does that by following the KRR principles in a Design Science Research methodological approach \cite{DSR, KRR}. By systematically formalising and encoding domain knowledge, e.g., spanning engagement scenarios, AI architecture components (data, models, rules, autonomy levels), and civilian assets, the assessment model ensures that representation and inference are developed responsibly throughout the research lifecycle. In addition, core concepts, interactions, and causal pathways linking AI engagement vectors to unintended effects on civilians and infrastructure are defined, modelled, and subjected to transparent reasoning processes. To capture the real-world complexity and uncertainty that military operations intrinsically have, the model incorporates collateral damage assessment metrics that capture temporal, spatial, force, severity, likelihood aspects, architectural vulnerability attributes, and context factors, thereby guaranteeing that assessments remain accountable, interpretable, and aligned with mission objectives and constraints. 

In this process, an upper-level taxonomy of engagement and collateral damage concepts is established in the context of engaging AI systems, then class hierarchies are elaborated with semantic relationships, axioms, and rule sets in a semantically rich and operationally grounded model. At the same time, an iterative, modular implementation guides the model's refinement: initial knowledge elicitation informs taxonomy construction, successive modeling cycles add constraints and rules, and the logical consistency and inferential fidelity verification through the reasoner is conducted. This assures a comprehensive assessment approach that lays a robust foundation for evidence-based collateral damage assessment and proportionality assessment decisions in diverse military operational contexts.

\section{Assessment Model\label{section:modeldesign}}
The proposed collateral damage assessment model is founded on the bedrock of the principle of distinction, whereby an AI system employed in an adversary's command, control, or critical infrastructure qualifies as a lawful military objective under international humanitarian law. By explicitly recognising the AI target's legitimacy, the model delineates clear boundaries between permissible attacks on hostile capabilities and prohibited harm to civilian persons and objects. This legal framing ensures that every engagement decision is conducted on a rigorous distinction analysis: only those AI systems whose neutralisation contributes to military advantage are considered for targeting, while others, and in particular, civilians and civilian objects are protected from attack. 

Building on this foundation, the systematically captures three fundamental dimensions of collateral effects, i.e., temporal, spatial, and force by encoding duration distributions (e.g., immediate versus sustained outages), propagation spreading (from the local data center to transnational network nodes), and effect types (service disruption, data corruption, physical destruction, etc.). At the same time, the model differentiates among data-driven, knowledge-driven, and neuro-symbolic architectures that AI systems have, mapping each component (datasets, inference engines, rule bases, autonomy modules, explainability interfaces) to potential failure modes and dependency chains. Furthermore, through a hybrid qualitative–quantitative schema, severity levels (ranging from negligible disruption to fatal or catastrophic destruction) are paired with probabilistic likelihood levels, enabling commanders to weigh both the intensity and the probability of unintended civilian impacts. This dual-mode perspective produces a detailed and auditable assessment ensuring that both numeric and linguistic aspects are represented and accounted in the reasoning process to provide responsible decision-making support to the military Commanders and their teams. Accordingly, the model is named CDAIMO (Collateral Damage Assessment from AI Engagement in Military Operations) is formally defined as follows:

\begin{figure}[htbp]
\centerline{\includegraphics [width=4cm]{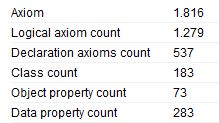}} 
\caption{Model metrics.}
\label{fig: cda_metrics}
\end{figure}

\begin{equation}
CDAAIMO = (C, A, R, I) 
\label{eq:defontology}
\end{equation}

where:
\begin{itemize}
    \item[] \textit{C} = the set of entities or classes that contain the core concepts of the model.
    \item[] \textit{A} = the set of attributes or characteristics of the concepts. 
    \item[] \textit{R} = the set of relationships between the instances or individuals of the concepts. 
    \item[] \textit{I} = the set of individuals, objects, or data values of the entities in domain.
\end{itemize}

\begin{figure}
\centerline{\includegraphics [width=3cm]{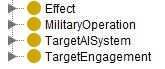}} 
\caption{Upper-classes of the model.}
\label{fig: cda_upperclass}
\end{figure}

\begin{figure}
\centerline{\includegraphics [width=8cm]{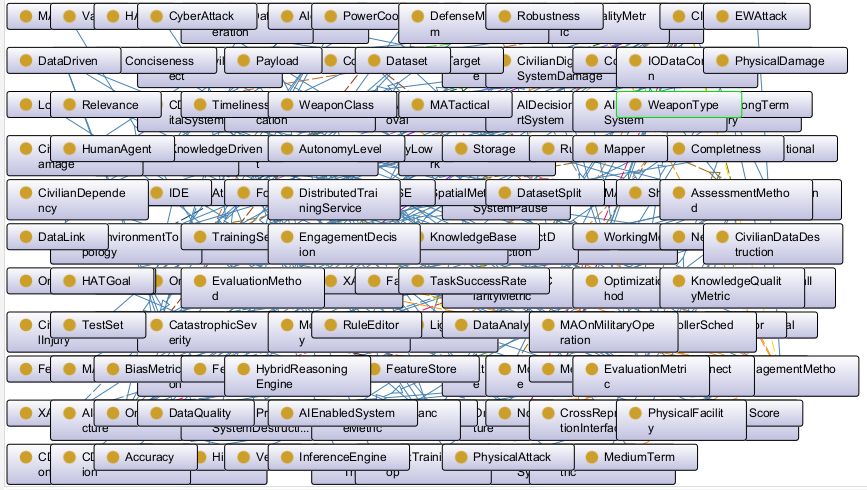}} 
\caption{More classes of the model.}
\label{fig: cda_moreclasses}
\end{figure}

The set \textit{C} embeds two types of classes. The first type is represented by upper-classes that provide a general understanding of the assessment scope, and the second type which is represented by the lower-classes which are sub-classes of the upper-classes and further detail each dimension considered in the assessment process. To this end, the upper-classes and a part of the sub-classes are discussed below and depicted in Figure~\ref{fig: cda_upperclass} and Figure~\ref{fig: cda_moreclasses}, respectively.

The class \textit{TargetAISystem} which captures information about the core types of AI systems (i.e., data-driven, knowledge-driven, and neuro-symbolic), categories of AI systems (i.e., AI-DSS, AI-Enabled Systems, and AI-Enabled Weapon Systems) together with architectural and functional elements (e.g., hyperparameter, accuracy, human-AI teaming context). Among its sub-classes are \textit{Dataset}, \textit{Precision}, \textit{Rule}, \textit{InferenceEngine}, and \textit{AutonomyLevel}.   

The class \textit{MilitaryOperation} which represents the context in which an AI system is engaged as a legitimate military target. At the same time, information about the method and metrics used for assessing the collateral damage expected from this engagement is considered through the temporal, spatial, and force dimensions together with severity and likelihood of occurrence of these unintended effects. To this end, among the sub-classes contained are recalled \textit{AssessmentMethod}, \textit{RuleQuality}, \textit{InterpretationClarity}, and \textit{OnTarget}.   

The class \textit{TargetEngagement} which contains information about the method, weapon, and attack vector that will be used in the engagement process. As sub-classes are mentioned here \textit{AttackVector}, \textit{EngagementDecision}, as potential engagement methods: \textit{CyberAttack}, \textit{EWAttack} and \textit{PhysicalAttack}.

The class \textit{Effect} which embeds the expected intended and unintended effects resulted from this engagement, i.e., military advantage and collateral damage, and their connection with the proportionality assessment. Given the scope of this model to assess collateral damage, further entities considered are e.g., \textit{CollateralDamageLevel}, \textit{CollateralDamageMitigationAction}, \textit{CivilianPhysicalInjury}, \textit{CivilianDataDestruction}, \textit{CivilianDigitalSystemDisruption}, and \textit{CollateralDamageTolerance}.

The set \textit{P} contains the attributes or data properties that the classes have in this context, by this characterising various dimensions that the instances or objects (\textit{I}) of these classes have through values such as string, integer, and double \textit{V}. This set is defined in equation~\ref{eq:dpontology}:

\begin{equation}
\mathcal{P} : \mathcal{I} \to \mathbb{V}
\label{eq:dpontology}
\end{equation}

\begin{figure}
\centerline{\includegraphics [width=4cm]{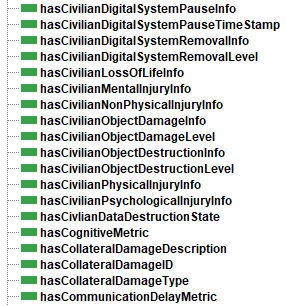}} 
\caption{Properties of the model.}
\label{fig: cda_dataprop}
\end{figure}

To this end, a part of the properties are illustrated in Figure~\ref{fig: cda_dataprop} and further presented: 
\begin{itemize}
    \item \textit{hasAccuracy} contains the accuracy value obtained through the evaluation of an AI model, and is of type \textit{double}.
    \item \textit{hasAITechnique} captures the type of AI technique used to implement the AI system, and is of type \textit{string}.
    \item \textit{hasDefenseMechanism} points out to the existence of a defense mechanism that would protect the AI system from attack, and is of type \textit{boolean}.
    \item \textit{hasAttackVectorID} embeds the identifier of a specific attack vector (e.g., malware, DDoS, jamming) that is used for target engagement, and is of type \textit{integer}.   
    \item \textit{hasCyberAttackStatus} shows the current engagement status that is being done using a cyber weapon as an attack vector, and can be either \textit{active, inactive} or \textit{disabled}.
    \item \textit{hasCDOnCivilianDigitalSystemInfo} has information about the civilian digital system affected by the engagement of an AI system, and is of type \textit{string}.
    \item \textit{hasCivilianDataAlterationLevel} contains information about the degree of alteration or damage that civilian data is experiencing through the attack, and can be one of the following values: \textit{very low, low, medium} and \textit{string}.
    \item \textit{hasConsistency} shows the degree of consistency that information has from the engagement process, applicable to both information or rule knowledge, and is of type \textit{double}.
    \item \textit{hasLongTermImpact} embeds the level of collateral damage expressed on a long term, and is of type \textit{double}.    
\end{itemize}

The set \textit{R} illustrates the relation between the individuals or objects of different classes of the model and defined in the equation~\ref{eq:defontology} as follows: 

\begin{equation}
\mathcal{R}_o : \mathcal{I} \times \mathcal{I} \to \mathbb{B}
\label{eq:opontology}
\end{equation}

where:
\begin{itemize}
    \item[] \textit{I} = the set of individuals or objects that classes of the model have
    \item[] \textit{B} = the set of boolean values \textit{true, false} which show if there is a belonging relation or not to a specific class.
\end{itemize} 

\begin{figure}
\centerline{\includegraphics [height=5cm]{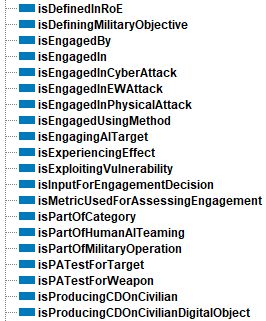}} 
\caption{Relationships of the model.}
\label{fig: cda_obprop}
\end{figure}

To this end, a selection of relationships is further presented and captured in Figure~\ref{fig: cda_obprop}:
\begin{itemize}
    \item \textit{isAssessedBy} captures the human decision-maker responsible for the collateral damage assessment decision. This is a relation between instances of the class \textit{AssessementDecision} and \textit{DecisionMaker}. 
    \item \textit{isUsingRoE} illustrates the fact that RoE considerations about military, legal, and political aspects that characterize the context of a military operation needs to be accounted. This is a relation between objects of the class \textit{MilitaryOperation} and \textbf{RoE}.  
    \item \textit{hasModelPerformance} shows the performance that an AI model has in relation to a specific task that is pursuing. This is a relation between objects of classes \textit{AITechnique, AISystemType, AISystemCategory} and \textit{ModelPerformance}.
    \item \textit{hasVulnerability} points out to an existing vulnerability that an AI system has. This is a relation between instances of classes \textit{AISystem} and \textit{Vulnerability}.
    \item \textit{isExploitingVulnerability} captures the relationship between an exploit developed to engage an existing vulnerability that an AI system has. This is a relation between objects of the classes \textit{Exploit} and \textit{Vulnerability}.
    \item \textit{isContributingToCollateralDamage} shows the fact that collateral damage is further produced through existing direct and indirect connections that the AI systems have with other military, dual-use, and civilian systems. This is a relation between instances of the classes \textit{Coonection} and \textit{CollateralDamage}.
    \item \textit{isMetricUsedForAssessingEngagement} depicts the assessment metrics used in this process as expected to produce collateral damage from engaging an AI system. This is a relation between instances of the classes such as \textit{SpatialMetric, TemporalMetric, ForceMetric, SeverityMetric} and \textit{EngagementMethod}. 
    \item \textit{isProducingEffect} captures in this context the sources of effects produced from engaging an AI system. This is a relation between objects of the classes \textit{TargetEngagement, MilitaryOperation} and \textit{Effect}. 
    \item \textit{hasTemporalAssessment} shows the temporal metrics (i.e., short term, medium term, long term) that are taking into account in the assessment process. This is a relation between instances of the classes \textit{TargetEngagement} and \textit{TemporalMetric}.    
\end{itemize}

Building upon the structured representation developed for collateral damage assessment, formalised rules are introduced to drive the reasoning process. These rules systematically interpret the interrelationships and attributes encoded within the assessment, enabling automated and consistent evaluation of potential collateral effects. For instance, rules may specify that if an AI system shares computational resources with civilian infrastructure, and the spatial assessment reveals a regional scope, then the likelihood and severity of service disruption to civilian assets must be adjusted accordingly. Furthermore, rule sets incorporate thresholds for temporal duration, severity, and probability, ensuring proportionality checks and legal compliance are rigorously enforced. By operationalising these rules within the assessment methodology, the model supports dynamic, scenario-driven reasoning that can adapt to diverse and complex settings in military operations targeting AI systems, thus providing transparent and justifiable recommendations for mitigating unintended civilian harm.

Rule 1: To capture the relation between an existing data vulnerability and potential collateral damage by defining any decision that is tied to an engagement against an AI system whose DataQualityMetric score does not exceed 0.5, and which also produces an instance of CollateralDamage, as a member of the class Effect which contains collateral damage risk.

\begin{align*}
&\textit{AssessmentDecision} \text{ and} \\
&\quad (\textit{isassessedBy some} ( \\
&\qquad \textit{TargetEngagement} \text{ and} \\
&\qquad\quad (\textit{hasTargetAISystem some} ( \\
&\qquad\qquad \textit{TargetAISystem} \text{ and} \\
&\qquad\qquad\quad \textit{isValidatedBy some} ( \\
&\qquad\qquad\qquad \textit{DataQualityMetric} \text{ and} \\
&\qquad\qquad\qquad (\textit{hasDataQuality max } 0.5)))) \\
&\qquad \text{ and } (\textit{isProducingEffect some CollateralDamage}))) \\
&\textit{SubClassOf Effect}
\end{align*}

Rule 2: To reduce collateral damage when both the probability of collateral damage is very high and the expected impact would be severe, then a flag is associated with the engagement and a trigger to mitigation decision is provided in order to ensure that the operation will consider collateral damage mitigation measures in order to be able to minimise the expected collateral risk on civilians and civilian objects. 

\begin{align*}
&\textit{AssessmentDecision} \text{ and} \\
&\quad (\textit{hasLikelihoodMetric some} ( \\
&\qquad \textit{LikelihoodMetric} \\
&\qquad \text{ and } (\textit{hasProbability min } 0.75))) \text{ and} \\
&\quad (\textit{hasSeverityMetric some} ( \\
&\qquad \textit{SeverityMetric} \\
&\qquad \text{ and } (\textit{hasSeverity value } \text{"Severe"}))) \\
&\textit{SubClassOf} \\
&\quad (\textit{hasAssessmentDecision some CDMitigationMethod})
\end{align*}

Rule 2 is fully auditable: one can trace exactly which severity and likelihood metrics triggered the decision, satisfying requirements for accountability and post-hoc review.

\section{Evaluation\label{section:modelevaluation}}
For evaluation purposes, a use case demonstration is conducted to reflect the model's effectiveness and support. In this virtual use case, a state actor identifies an adversarial AI-DSS operating within a hostile C2 (Command and Control) network. A Cyber Operation is prepared to degrade its hostile functionality. The system is considered under the \textit{TargetAISystem} class as an \text{AI-DSS} sub-class that integrates an AI data-driven architecture type. Its components include a real-time inference engine and proprietary civilian-based datasets. The \textit{MilitaryOperation} class frames this cyber engagement within lawful boundaries of armed conflict, emphasising adherence to principles such as distinction and proportionality. The attack vector is instantiated from the \textit{CyberAttack} sub-class of \textit{TargetEngagement}, using a malware payload with \textit{hasAttackVectorID} = 1002 targeting a 0-day software vulnerability (\textit{hasVulnerability}). A \textit{TemporalMetric} with short-term duration and a \textit{ForceMetric} indicating non-kinetic disruption are used to project potential collateral damage, such as downstream service outages in civilian hospitals using the same data infrastructure. This information is captured using the \textit{isProducingEffect, hasTemporalAssessment} and \textit{isContributingToCollateralDamage} relationships. Collateral damage is anticipated due to the shared computational backbone between the military AI system and civilian systems, e.g., public emergency response coordination platforms. The \textit{Effect} class is populated with instances such as \text{CivilianDigitalSystemDisruption} and \textit{CivilianDataDestruction}, indicating possible loss of emergency services data and degraded civilian coordination on services. The data property \textit{hasCivilianDataAlterationLevel} is assessed to be \textit{high}, indicating substantial risk.

Through rule 1, collateral risk classification is considered under the \textit{Effect} class if the \textit{DataQualityMetric} of the AI system is under 0.5, showing unreliable partitioning of civilian versus military datasets. Due to poor data labeling in the adversary's infrastructure, a score of 0.45 is registered, validating the established conditions. Consequently, this decision is subclassed as an instance of \textit{Effect} containing collateral damage and an alert is provided to the \textit{DecisionMaker} through \textit{isAssessedBy} to initiate mitigation evaluation before engagement. By recognising the potential civilian impact severity (\textit{hasSeverity} = Severe) and likelihood of collateral damage (\textit{hasProbability = 0.81}), through rule 2 the \textit{hasAssessmentDecision} points to a \textit{CDMitigationMethod}, such as delaying malware deployment until civilian systems can be temporarily decoupled. This ensures engagement adheres to legal and ethical norms while maintaining operational effectiveness. Through this use case, the role, approach, and contribution of the proposed model is illustrated.

\section{Conclusion\label{section:conclusion}}
This research introduces a computational collateral damage assessment model for AI system targeting that captures temporal, spatial, and force dimensions through KRR principles and DSR methodology. The model integrates qualitative and quantitative assessment perspectives while accounting for different AI system types and contextual engagement factors. Its transparent, modular architecture ensures auditability and interoperability. Future research will incorporate LLM-driven scenario generation and RL-based optimization to refine assessments and mitigate risks, enabling experimentation across multiple warfare domains for responsible military AI development.

\bibliographystyle{IEEEtran}
\bibliography{references}
\vspace{12pt}

\end{document}